\documentclass[runningheads]{llncs}

\usepackage[acronym]{glossaries}
\usepackage{acronyms}
\usepackage[export]{adjustbox}
\usepackage{array,makecell}
\newcolumntype{P}[1]{>{\centering\arraybackslash}p{#1}}
\usepackage{multirow,tabularx}
\usepackage[polish, english]{babel}

\usepackage[T1]{fontenc}
\usepackage{graphicx}
\usepackage{comment}
\usepackage{amsmath}
\usepackage{color}
\usepackage{url}
\usepackage{hyperref}
\hypersetup{
	 hidelinks = true,
   colorlinks = true,
   citecolor = blue,
   urlcolor = blue,
   linkcolor = blue
}

\usepackage{booktabs}
\usepackage[capitalize]{cleveref}
\crefname{figure}{Fig.}{figures}
\Crefname{figure}{Figure}{Figures}
\crefname{table}{Tbl.}{tables}
\Crefname{table}{Table}{tables}
\crefname{section}{Sect.}{sections}
\Crefname{section}{Section}{Sections}

\usepackage{soul}
\usepackage[usenames,dvipsnames]{xcolor}

\newif\ifreview
\reviewfalse

\begin{document}
\def\GCPRTrack{Regular Paper}

\title{Uncertainty Driven Active Learning for Image Segmentation in Underwater Inspection\thanks{Partially supported by the European Union’s Horizon 2020 research and innovation programme under the Marie Skłodowska-Curie grant agreement No 956200, REMARO}}
\titlerunning{Active Learning for Underwater Inspection}
%

\ifreview
	

    \authorrunning{ROBOVIS 2024 Submission. CONFIDENTIAL REVIEW COPY.}
	\author{ROBOVIS 2024}
    \institute{\GCPRTrack{}}
\else
    \author{Luiza Ribeiro Marnet\inst{1,2}\orcidID{0000-0001-6717-9306} \and \\
    Yury Brodskiy\inst{1}\orcidID{0009-0002-0445-8126} \and
    Stella Grasshof\inst{2}\orcidID{0000-0002-6791-7425} \and
    Andrzej Wąsowski \inst{2}\orcidID{0000-0003-0532-2685}}
    \authorrunning{L. Ribeiro Marnet et al.}
    %
    \institute{EIVA a/s, Denmark \\
    \email{\{lrm,ybr\}@eiva.com}\\
    \and
    IT University of Copenhagen, Denmark\\
    \email{\{stgr,wasowski\}@itu.dk} \\
    }
\fi
\maketitle              
\begin{abstract}
Active learning aims to select the minimum amount of data to train a model that performs similarly to a model trained with the entire dataset.  We study the potential of active learning for image segmentation in underwater infrastructure inspection tasks, where large amounts of data are typically collected. The pipeline inspection images are usually semantically repetitive but with great variations in quality. We use mutual information as the acquisition function, calculated using Monte Carlo dropout. To assess the framework’s effectiveness, DenseNet and HyperSeg are trained with the CamVid dataset using active learning. In addition, HyperSeg is trained with a pipeline inspection dataset of over 50,000 images. For the pipeline dataset, HyperSeg with active learning achieved 67.5\% meanIoU using 12.5\% of the data, and 61.4\% with the same amount of randomly selected images. This shows that using active learning for segmentation models in underwater inspection tasks can lower the cost significantly. 

\end{abstract}

\section{Introduction}
\label{Sec:introduction}

\noindent
Computer vision plays a pivotal role in advancing automation across various applications, such as equipment inspection~\cite{wang2020smart,bouarfa2020towards,guo2021automatic}, autonomous driving~\cite{xiao2023baseg,chen2017multi,7780605}, medical diagnoses~\cite{zhou2020cnn,polsinelli2020light,desai2021anatomization}, underwater debris detection~\cite{DBLP:journals/corr/abs-2007-08097,9882484}, and underwater pipeline inspection~\cite{gavsparovic2022deep,8310098}. However, the large amounts of annotated datasets required for training these models presents a major challenge. Annotating such datasets is time-consuming, expensive, or infeasible, especially in domains requiring expert knowledge.

Even though access to large data is important, the quality of the data is critical. Active learning focuses on selecting the smallest sample set that can be used to train a model to achieve the same performance as when training with the entire dataset\,\cite{ren2021survey,budd2021survey}. This typically involves training the model with few initial samples and selecting additional samples for labeling and retraining the model iteratively\,\cite{budd2021survey}. It is known that random selection usually does not perform well.
\looseness -1

Epistemic uncertainty~\cite{abdar2021review} can be used to identify samples that are most informative for training deep learning models. It measures the model's confidence in its predictions based on its level of familiarity with the input data~\cite{abdar2021review,NIPS2017_2650d608}. By identifying samples that induce high epistemic uncertainty, the most informative samples can be labeled and added to the training dataset, potentially reducing the labeling effort while improving the model's performance. 

\begin{figure}[t!] 
    \centering

    \includegraphics[width=0.91\linewidth]{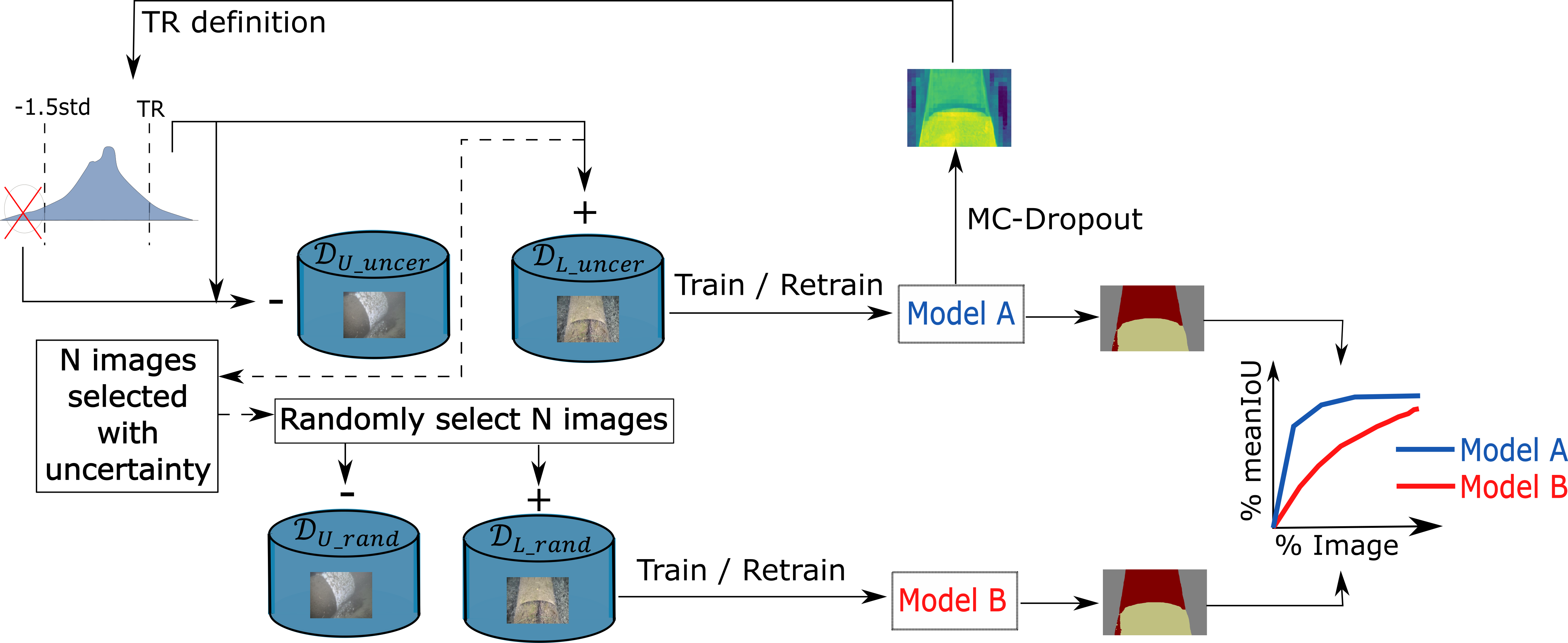}

    \caption{Our active learning process: After pre-training the model with a small set of randomly chosen images, we start selecting images with uncertainty above a threshold \(\textrm{TR}\) (top-left) for training and validation of Model A. Due to space constrains only one $TR$ is represented. However, in each iteration two new thresholds are defined: $TR_t$ for training, and $TR_v$ for validation images. We discard images with an uncertainty below the mean minus 1.5 standard deviations. For evaluation, the process is repeated on randomly selected images (bottom-left) training model B.  We compare the meanIoU performance of the two models after each iteration (bottom-right). In this image, $D_U$ and $D_L$ are the pools of unlabeled and labeled images, respectively, both for uncertainty-based selection ($uncer$) and random selection ($rand$). The signs '$+$' and '$-$' indicate that the newly selected images are removed from $D_U$ and added to $D_L$. The arrows with dashed lines highlight that, for each iteration, the number of images to be selected randomly is defined by the number of images selected based on uncertainty. \looseness -1}

    \label{fig:images_selection}%

    \vspace{-3mm}
\end{figure}

In this paper, we investigate the use of epistemic uncertainty for image selection in the field of computer vision applications, cf.\, \cref{fig:images_selection}, more specifically to enhance the capabilities of autonomous robotic systems. We employ two distinct datasets, one for street view image segmentation, a crucial component in the domain of autonomous vehicle navigation, and the other for segmenting images captured during underwater robotics missions. Segmenting underwater images is a challenging task that can aid in, e.g., underwater autonomous vehicle path tracking, and is especially important for visual inspection of underwater equipment and structures. Our approach is validated using real underwater images from various missions and locations. We demonstrate the performance of the active learning method with epistemic uncertainty using two models, DenseNet for semantic segmentation~\cite{jegou2017one} and HyperSeg~\cite{nirkin2021hyperseg}, and two datasets, CamVid~\cite{BrostowSFC:ECCV08,brostow2009semantic} and real RGB images from underwater surveys.

The results for the underwater images are specially important. These images come from a huge pool of images for pipeline inspection, and are semantically similar, but vary in quality due to factors like low and non-uniform illumination, color degradation, low contrast, blurring, and hazing~\cite{schettini_underwater_2010,duarte_dataset_2016,chiang_underwater_2012}. Labeling these images requires specialized expertise and constitutes a significant cost for the growing but still small underwater automation industry. Selecting key images representing the different image qualities,  reduces the required amount of training data, resulting in lower cost and human effort for labeling, which is valuable for innovating companies in this sector.

To the best of our knowledge, this is the first paper applying active learning for real underwater images. Our contributions include:
\begin{itemize}
    \item A systematic study of the method against a random selection baseline, showing 6.2\% gain in the meanIoU over the baseline in the underwater dataset.
    \item An active learning framework that uses a threshold for selecting new images instead of the usual fixed percentage of images at each active learning cycle.
    \item An implementation and reproduction package allowing to reproduce the results on the CamVid dataset using DenseNet and HyperSeg (the underwater imagery is unfortunately security-sensitive and cannot be released).
\end{itemize}

\section{Related Work}
\label{Sec:related_work}

\noindent


\noindent
In active learning with classical machine learning methods, it is common to select new samples using a metric for capturing uncertainty and another for measuring similarity. The former identifies samples for which the model is unsure about the predicted output, and is used for selecting the most informative samples for the model to learn from. The last aids in selecting representative samples and preventing the selection of redundant samples with similar information. 


Thus, the first step in developing an active learning framework is to decide on an uncertainty metric. While the softmax values are often used as a measure that reflects the model's confidence for classification tasks, it has been demonstrated that misclassified samples can have high softmax values\,\cite{oberdiek2018classification}, for example, when an input out of the distribution of the training dataset is used during inference\,\cite{gal2016uncertainty}. Therefore, other methods of capturing uncertainty should be used.

The uncertainty of the predictions, called predictive uncertainty, can be decomposed into epistemic and aleatoric\,\cite{abdar2021review}. Epistemic uncertainty reflects the model's lack of knowledge about the input's pattern, while aleatoric uncertainty reflects the data quality itself, e.g. noise caused by the sensor that captures the data. Since the goal of active learning is to select new images that bring knowledge that the model lacks for training, the important uncertainty in this scenario is the epistemic. Methods like \gls{MC-Dropout}, deep ensembles, and error propagation can be used to access this uncertainty\,\cite{feng2021review}.



\Gls{MC-Dropout} is a widely used method for modeling epistemic uncertainty in deep neural networks. During inference dropout layers are used and the same input sample is passed forward multiple times through the model. Since dropout is applied, each pass may produce a different output, and these outputs are used to assess the epistemic uncertainty. If the model is well-trained and the input is similar to what the model has seen during training, the outputs will be similar or identical. On the other hand, if the model does not have enough knowledge about an input, the outputs of each forward pass will be more varied, indicating high uncertainty. Different metrics, such as variation ratio, mutual information, total variance, margin of confidence, and predictive entropy, can be used for that~\cite{milanes2021monte}.

Ensembles of networks can also be used to predict epistemic uncertainty. In this method, an input sample is passed through the model and the results of each network in the ensemble are used to estimate the uncertainty\,\cite{9353390}. Finally, the error propagation method\,\cite{postels2019sampling} estimates the model uncertainty by propagating the variance of each layer to the output. This variance arises in layers such as dropout and batch normalization and is modified by the other layers of the model.
\looseness -1

Deep active learning has been applied to tasks requiring costly labeling, such as medical image analysis. Softmax confidence with a single forward pass was used to evaluate the segmentation uncertainty for pulmonary nodules~\cite{wang2019nodule} and membrane~\cite{7532697} images, even if these may not be ideal for deep learning models.
Other studies have proposed more reliable approaches. Using different subsets of training samples of biomedical images, a set of segmentation models was trained, and the uncertainty was measured as the variance between their outputs~\cite{yang2017suggestive}. Different metrics, such as max-entropy and \gls{BALD}, were calculated using \Gls{MC-Dropout} outputs for selecting new images to label for training skin cancer image classifier\,\cite{gal2017deep} and to segment medical images~\cite{saidu2021active}. Moreover, a comparison between querying entire medical images and querying image paths concluded that the latter led to better models\,\cite{li2020uncertainty}.

More recently, active learning was studied in the autonomous driving context. The Deeplabv3+ architecture with a Mobilenetv2 backbone was trained with uncertainty- and difficulty-driven image selection\,\cite{xie2020deal}. A further reduction of labeled pixels was obtained by querying image paths\,\cite{sreenivasaiah2021meal,rangnekar2023semantic}. Even though these studies used entropy-based metrics for image selection, the metrics were calculated using a single pass softmax output and did not use MC-dropout.
\looseness -1

In this work, we use MC-dropout to guide active learning in real-world problems. We first validate the method with DenseNet and HyperSeg for semantic segmentation\,\cite{jegou2017one} on the publicly available street view dataset CamVid\,\cite{BrostowSFC:ECCV08,brostow2009semantic}. We then use active learning with the better-performing model, HyperSeg~\cite{nirkin2021hyperseg}, to obtain a new state-of-the-art model for real-time semantic segmentation in an underwater application. Our underwater dataset contains many images that are hard to segment even for human eyes. The dataset is unbalanced, with some classes appearing in only a few images and occupying a low percentage of pixels. The goal is to analyze the effectiveness of active learning in training the models with only a small percentage of the data and in learning to predict underrepresented classes.
\looseness -1

\section{Method}%
\label{Sec:methodology}

\noindent
We now present an overview of our learning method:  the models, the acquisition function, and the entire active learning framework.

\paragraph{Models.}

We employ two deep learning models for segmentation, DenseNet and HyperSeg. DenseNet~\cite{Huang_2017_CVPR}, initially developed for classification, was later extended for segmentation~\cite{jegou2017one}. We choose DenseNet due to its success in accessing epistemic uncertainty using MC-Dropout~\cite{NIPS2017_2650d608}. We utilize the lighter version, DenseNet-56.

HyperSeg is a state-of-the-art model based on the U-Net architecture\,\cite{nirkin2021hyperseg}. In the decoder, it utilizes dynamic weights, that are generated based on both the input image and the spatial location. Because of its high \gls{meanIoU}, high \gls{FPS} rate, and the option to use dropout, we hypothesize that it is an appropriate choice. We use the version HyperSeg-S with efficientNet-B1 as the backbone\,\cite{tan2019efficientnet}, as it has achieved \gls{FPS} of 38.0 and \gls{meanIoU} of 78.4\% in the original work. To the best of our knowledge HyperSeg has not been used in similar experiments to this date.
\looseness -1

\paragraph{Acquisition Function.}

We calculate the epistemic uncertainty using \gls{MC-Dropout} with \textit{T} forward passes. The metric used was the mutual information, that for a segmentation problem with $\mathcal{C}$ classes is calculated for each pixel of the image as:  
\begin{equation} \label{eq:mutual_information}
    \mathcal{I} ~ = ~ \underbrace{ \frac{-1}{\log_2(C)}\sum_{c=1}^{C}p_c^*\log_2(p_c^*) }_{\mathcal{H}} \quad + \quad \frac{1}{T\log_2(C)}  \sum_{c=1}^{C}\sum_{t=1}^{T}p_{ct}\log_2(p_{ct}),
\end{equation}
where $\mathcal{H}$ is the entropy, %
\noindent
$p_t = (p_{1t}, ..., p_{Ct})$ is the softmax output for a single forward pass $t$, and $p^* = (p_{1}^*, ..., p_{C}^*)$ is the average prediction of the \textit{T} passes. 
For each pixel, we predict one value $p^*$, and a class label $c$ based on the maximum value of $p_c^*$.
We divide the equation by $\log_2(C)$ to normalize the entropy and the mutual information between zero and 1, and -1 and 1, respectively.



\paragraph{Deep Active Learning Framework.}


In a real-world scenario the images are labeled after being queried. To reflect this, we only use the labels after the images are selected. Our active learning process begins by randomly selecting a constant \textit{P}\% of the images from both the training and validation sets in the first iteration. The model is then trained with these images. The epistemic uncertainty of each selected image is then calculated as the average uncertainty of each pixel:
\begin{equation} \label{eq:total_uncertainty_per_image}
  EU_{\textrm{img}} = \frac{1}{N}\sum_{i=1}^{N}EU_{i}, 
\end{equation}
where $EU_{i}$ is the epistemic uncertainty for pixel $i$, \cref{eq:mutual_information}, of an image with $N$ pixels. 
The next step is to calculate the thresholds $\textrm{TR}_{t}$ and $\textrm{TR}_{v}$ (\cref{fig:images_selection}) used to decide which images from the unlabeled training and validation datasets, respectively, should be selected for labeling. These thresholds are calculated as:
\begin{equation} \label{eq:tr}
    \textrm{TR}_{} = \overline{EU}_{\textrm{img}} + S\sigma
\end{equation}
where $\overline{EU}_{\textrm{img}}$ and $\sigma$ are the mean and standard deviation of $EU_{\textrm{img}}$, for the corresponding training and validation datasets, and $S$ is a positive constant defined by the user. Bigger values of $S$ result in querying fewer images. For the next iteration, it is computed the $EU_{\textrm{img}}$ of the images in the pools of unlabeled (remaining) training and validation data. The images with $EU_{\textrm{img}}$ above the respective thresholds, $\textrm{TR}_{t}$ and $\textrm{TR}_{v}$, are selected for labeling. The model is then retrained with the new and previously selected images.

Because the pipeline dataset is huge, analyzing it entirely per iteration is time-consuming.  
First, the images are shuffled, then $EU_{\textrm{img}}$ is calculated for each image. The image is selected or not based on the thresholds. 
The iteration stops when a pre-defined number of images is reached or the whole dataset is evaluated. 

To accelerate the selection process, images with uncertainty values below 1.5 standard deviations below the mean are excluded from future selection. Since the model's predictions for these samples are relatively certain, using them to retrain the model is unlikely to improve performance.

For evaluation purposes, a second model is trained using randomly selected images with the same number of selected images as the uncertainty-based selection. The same initial set of images selected in the first iteration for the uncertainty study is used for the random selection, ensuring a fair comparison.

After each iteration, the models are evaluated using the test dataset, which remains the same in all iterations. A single model is trained during the first iteration and saved as two separate models, \textit{A} and \textit{B}, cf.\,\cref{fig:images_selection}. From the second iteration onwards, models \textit{A} and \textit{B} are trained separately, with \textit{A} using images selected based on uncertainty and \textit{B} using randomly selected images. The active learning process is iteratively repeated until a chosen condition is met, e.g. if all images have been selected, no improvement in \gls{meanIoU} on the test set is achieved, or few images have an epistemic uncertainty above a threshold. 
In this paper, for the CamVid experiments, we stop if very few images with uncertainty above the threshold are left. To observe if the performance of the models would suddenly improve, we keep the experiments running 
after unlabeled images stopped having uncertainty above the threshold. For the pipeline dataset, which is much more time-consuming, we stop if the meanIoU does not improve significantly anymore.

\paragraph{Training Details.}

DenseNet~\cite{jegou2017one} was implemented from scratch and trained following the original paper, which includes pre-training and fine-tuning phases. The model was initialized with HeUniform in the first iteration, and with the weights of the best model from the previous iteration in subsequent iterations. RMSProp optimizer was used with a weight decay of $1e^{-4}$, and horizontal flip on the training and validation datasets. Pre-training was performed using random crops of 224x224. The initial learning rate was $1e^{-3}$ in the first iteration and $1e^{-4}$ in the subsequent ones, with a learning rate decay of $1e^{-4}$. Fine-tuning was performed using 360x480 image resolution, an initial learning rate of $1e^{-4}$, and no learning rate decay. In each iteration, the training process stopped after no improvement in validation meanIoU or validation loss for 150 consecutive epochs, and the fine-tuning process stopped after no improvement for 50 consecutive epochs.
\looseness -1


For HyperSeg, we used the source code of the original authors.\footnote{\url{https://github.com/YuvalNirkin/hyperseg}} The model's backbone was initialized with imagenet pre-trained weights in the first iteration and with the weights of the best model from the previous iteration in subsequent iterations. The initial learning rate was 0.001 in the first iteration and 0.00098 in subsequent iterations. The training of each iteration stopped after no improvement in validation meanIoU or validation loss for 10 consecutive epochs. The version chosen, HyperSeg-S, operates on 768x576 resolution. 

Both models used a rate of 20\% in each dropout layer utilized. DenseNet has a dropout layer after each convolutional layer. Hyperseg can add dropout layers at the end of the encoder and at the end of the hyper patch-wise convolution blocks in the decoder. For HyperSeg, we only used the dropout layer in the encoder. All experiments were developed using PyTorch.
\looseness -1




\section{Results}
\label{Sec:results}

We discuss the optimal number of forward passes $T$ used in MC-dropout, and the impact of applying active learning on the datasets \emph{CamVid}~\cite{BrostowSFC:ECCV08,brostow2009semantic}, and \emph{pipeline}.

\paragraph{Datasets.}

CamVid \cite{BrostowSFC:ECCV08,brostow2009semantic} is a street view dataset for segmentation. It contains video frames captured from the viewpoint of a driving car. 
The original dataset provides 369 images in the training subset, 100 in the validation subset, and 232 in the testing subset, in a resolution of $720\times960$. 
The half-resolution version consists of 367 images for training, 101 for validation, and 233 for testing. 
Labels for 32 hierarchical classes are provided from which we use 11: `sky', `building', `column-pole', `road,' `sidewalk', `tree', `sign-symbol', `fence', `car', `pedestrian', and `bicyclist.'
\looseness -1

\begin{figure*}[!t]
  \centering

    \begin{tabular}{@{}c@{\hspace{0.3mm}}c@{\hspace{0.3mm}}c@{\hspace{0.3mm}}c@{\hspace{0.3mm}}c@{\hspace{0.3mm}}c@{}}

      \adjustbox{valign=m,vspace=.05mm}{\includegraphics[width=0.16\textwidth]{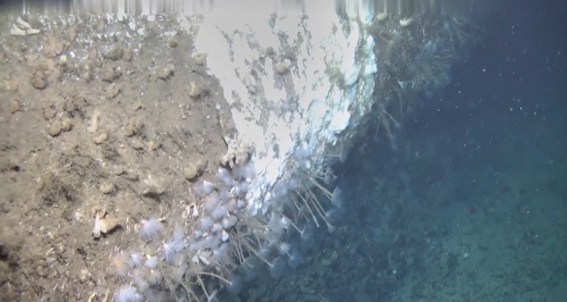}} &
      \adjustbox{valign=m,vspace=.05mm}{\includegraphics[width=0.16\textwidth]{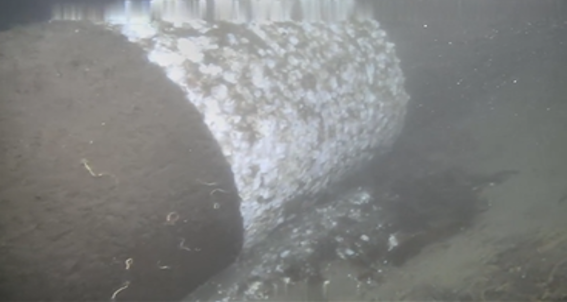}} &
      \adjustbox{valign=m,vspace=.05mm}{\includegraphics[width=0.16\textwidth]{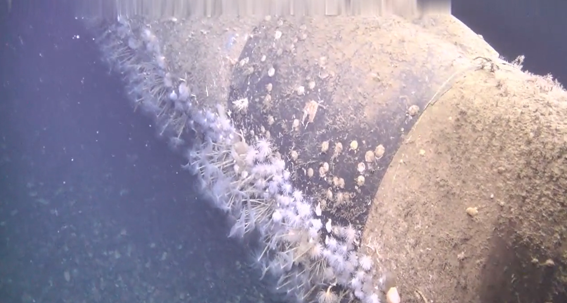}} &
      \adjustbox{valign=m,vspace=.05mm}{\includegraphics[width=0.16\textwidth]{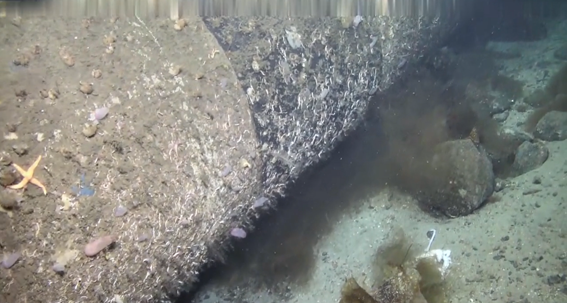}} &
      \adjustbox{valign=m,vspace=.05mm}{\includegraphics[width=0.16\textwidth]{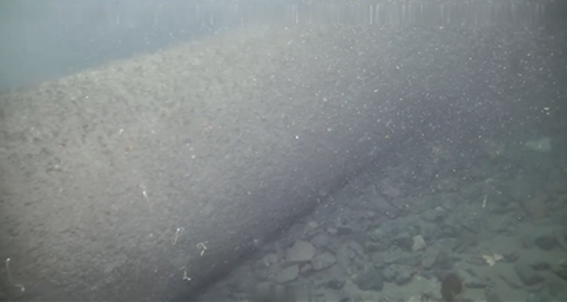}} &
      \adjustbox{valign=m,vspace=.05mm}{\includegraphics[width=0.16\textwidth]{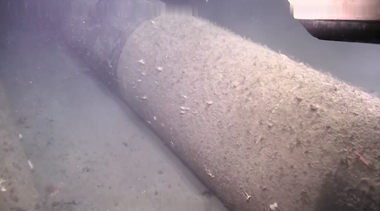}}
      \\[2.mm]

      \scriptsize (a)&\scriptsize (b)&\scriptsize (c)&\scriptsize (d)&\scriptsize (e)&\scriptsize (f)

  \end{tabular}

  \vspace{-1mm}

  \caption{Example of the pipeline dataset: (a-b) an anode and pipeline, (c) a pipeline and a field joint, (d) a field joint, boulders and pipeline, (e) a pipeline alone, and (f) a pipeline, field joint,  and part of a vehicle in the top right (best viewed in color).\label{fig:pipelinedataset}}

  \vspace{-2mm}
\end{figure*}

The underwater dataset \emph{pipeline} was provided by our industrial partner's customers and contains images from pipeline surveys. Due to security reasons,\footnote{E.g.\,\url{https://en.wikipedia.org/wiki/2022_Nord_Stream_pipeline_sabotage}} the data is not publicly available. 
The entire dataset comprises 64,920 video frames recorded during surveys of infrastructure. Out of these frames, we reserved 8,896 for testing while the remaining 56,024 images have been randomly split into 70\% for training and 30\% for validation. The dataset has five classes: `background', `pipeline', `field joint', `anode', and `boulder-and-survey vehicle', with respectively 99.9\%, 77.1\%, 20.3\%, 21.0\% and 15.5\% of the training and validation images in each class. \Cref{fig:pipelinedataset} shows several examples. In the figure, some images have a more brownish background and others are more blueish. Image (e) suffers from higher turbidity than the other images. Marine flora grows on the pipelines in images (a), (c), and (d). Underwater images suffer from low and non-uniform illumination, color degradation, low contrast, blurring, and hazing\,\cite{schettini_underwater_2010,duarte_dataset_2016,chiang_underwater_2012}. The poor quality is due both to light attenuation and floating particles in marine environments, known as ``marine snow''\,\cite{schettini_underwater_2010}. This causes haziness\,\cite{duarte_dataset_2016} and reduces visibility to below 20m, sometimes even below 5m. Also, the light is attenuated differently depending on the wavelength.  The red light is attenuated faster, while blue is attenuated slower\,\cite{chiang_underwater_2012} resulting in the blueish hue of images.\looseness -1


\begin{figure}[tb]
    \centering
    \includegraphics[width=0.85\linewidth]{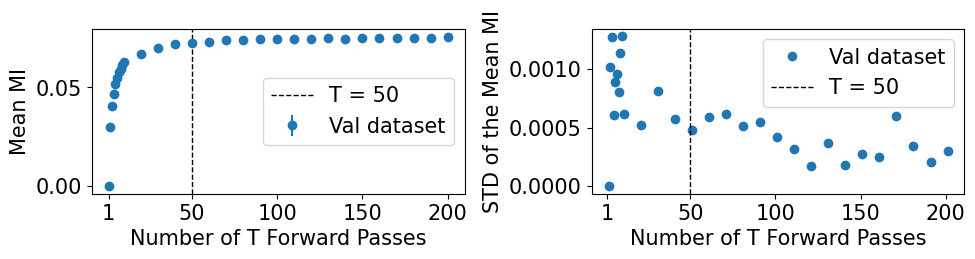}

    \caption{
    For each number of forward passes $T$, $\overline{EU}_{\textrm{img}}$ of the validation dataset of CamVid was calculated five times using DenseNet. 
    The graph on the left shows the average of the five results obtained, and the graph on the right shows the standard deviation. Here $\overline{EU}_{\textrm{img}}$ was calculated using mutual information (MI).}%
    \label{fig:Tpasses_camvid_mi}

    \vspace{-2mm}
\end{figure}

\paragraph{The Number of Forward Passes.}
In many papers the number of forward passes in MC-Dropout is set to \(T=50\), without clear justification. 
To address this shortcoming, we conducted a study to determine the optimal value of~\(T\). Since the epistemic uncertainty per image $EU_\textrm{img}$, \cref{eq:total_uncertainty_per_image}, is used for querying new images for being labeled, this study should focus on how the number of passes \(T\) affects \(EU_\textrm{img}\). 
%
%
We calculated $\overline{EU}_{\textrm{img}}$ of the dataset for $T = \{1, 2, 3, ..., 10\}$ and $T = \{20, 30, 40, ..., 200\}$. Each value of $T$ was evaluated five times. The study was conducted using DenseNet-56 trained and validated with the complete subsets of CamVid.  \Cref{fig:Tpasses_camvid_mi} presents the results for the validation dataset. The graphs illustrate the mean and standard deviation of the results for the five repetitions. The graphs indicate that a stable mean value is achieved after \textit{T} reaches 50, confirming the choice in published works. Furthermore, the standard deviation is over 100 times smaller than the mean, suggesting that 50 forward passes are enough to obtain a stable result with MC-Dropout. Henceforth in this paper, we define \textit{T} as 50, in line with other works.


\paragraph{Active Learning for the CamVid Dataset.}

\begin{figure}[tb]
    \centering
    \includegraphics[
      width = 1. \linewidth,
      clip,
      trim = 0mm 2.5mm 0mm 2.5mm
    ]{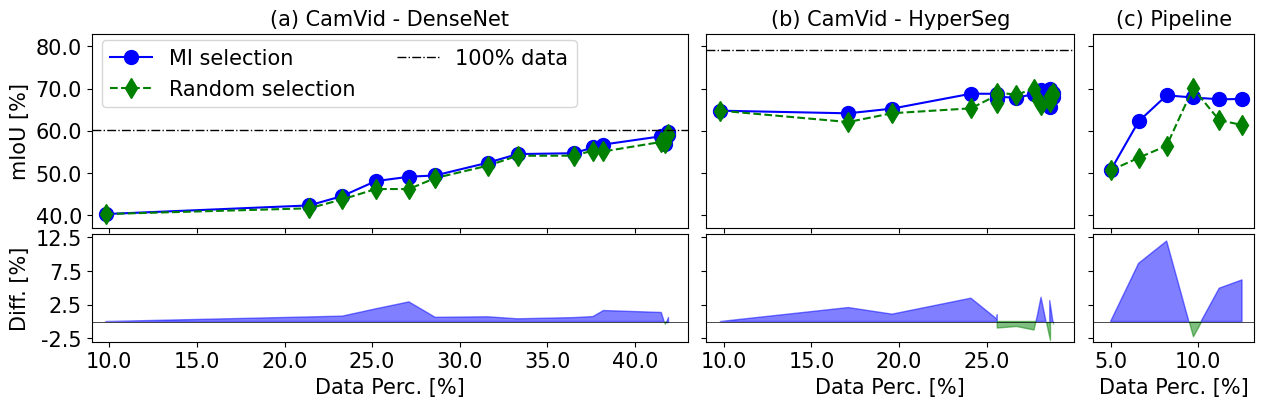}

    \vspace{-1mm}

    \caption{Results of the active learning experiments. The bottom graphs show the difference in mean IoU between the model trained with active learning vs. trained with random images, where positive values means active learning prevails.}
    \label{fig:camvid_pipeline_mi}

    \vspace{-2mm}
\end{figure}

\Cref{fig:camvid_pipeline_mi}(a) presents the outcomes of the experiment on the CamVid dataset using DenseNet, starting the process with $P=10\%$ images and $S=1.0$. 
The model's meanIoU stabilizes around 59\% after iteration 12, using approximately 40\% of the training data and slightly over 50\% of the validation data, which represents around 41\% of the data when the weighted mean is calculated for both subsets as shown in Figure \ref{fig:camvid_pipeline_mi}(a). After that, only one new image was selected for training and one for validation, indicating that additional images do not contribute significantly to the model's knowledge. Comparatively, training the model with the entire dataset yielded a slightly higher meanIoU of 60.22\% in our implementation. However, the model trained with uncertainty-selected images consistently outperformed the one trained with random images.
\looseness-1






\begin{figure}[t!]
  \centering

    \begin{tabular}{@{}c@{\hspace{0.3mm}}c@{\hspace{0.3mm}}c@{\hspace{0.3mm}}c@{\hspace{0.3mm}}c@{\hspace{0.3mm}}c@{\hspace{0.3mm}}c@{}}

      &
      \scriptsize Original &
      \scriptsize GT &
      \scriptsize 1 Pass &
      \scriptsize MCD &
      \scriptsize Entropy &
      \scriptsize MI \\

      \multicolumn{1}{c}{\rotatebox[origin=c]{90}{\scriptsize  DenseNet}}  & 
      \adjustbox{valign=m,vspace=.05mm}{\includegraphics[width=0.16\textwidth]{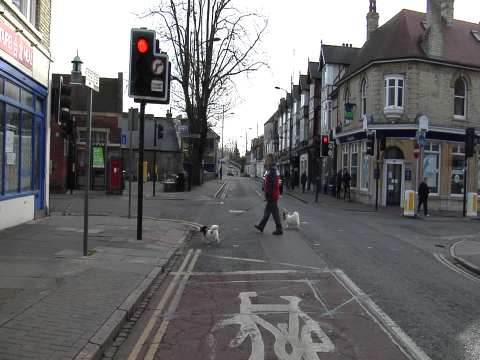}} &
      \adjustbox{valign=m,vspace=.05mm}{\includegraphics[width=0.16\textwidth]{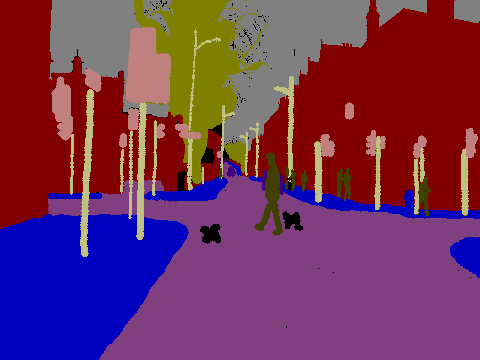}} &
      \adjustbox{valign=m,vspace=.05mm}{\includegraphics[width=0.16\textwidth]{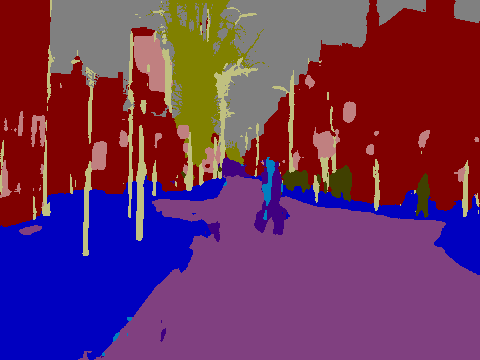}} &
      \adjustbox{valign=m,vspace=.05mm}{\includegraphics[width=0.16\textwidth]{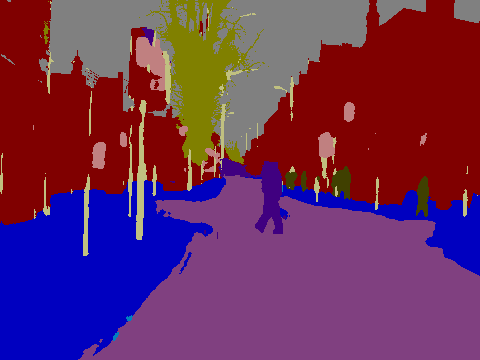}} &
      \adjustbox{valign=m,vspace=.05mm}{\includegraphics[width=0.16\textwidth]{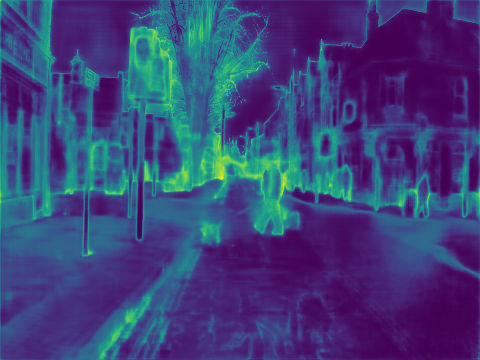}} &
      \adjustbox{valign=m,vspace=.05mm}{\includegraphics[width=0.16\textwidth]{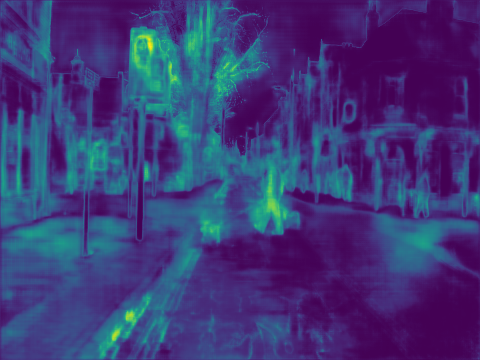}}
      \\[2.mm]

      \multicolumn{1}{c}{\rotatebox[origin=c]{90}{\scriptsize  HyperSeg}}  & 
      \adjustbox{valign=m,vspace=.05mm}{\includegraphics[width=0.16\textwidth]{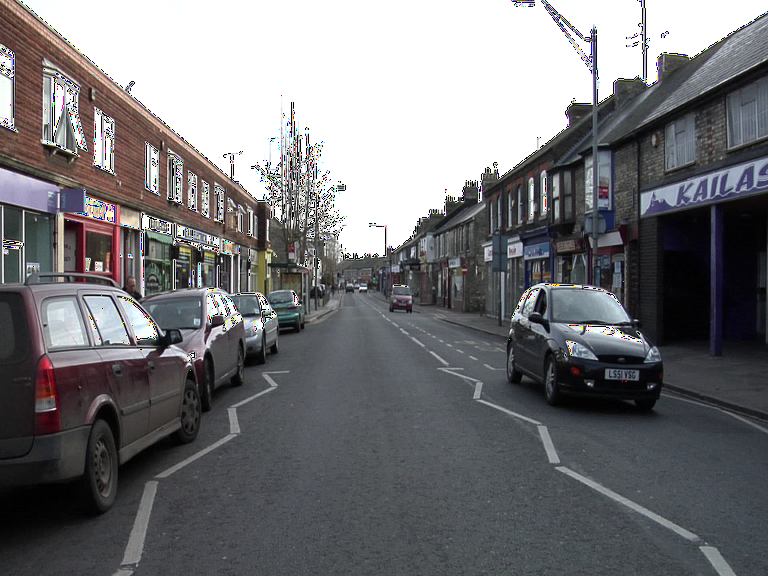}} &
      \adjustbox{valign=m,vspace=.05mm}{\includegraphics[width=0.16\textwidth]{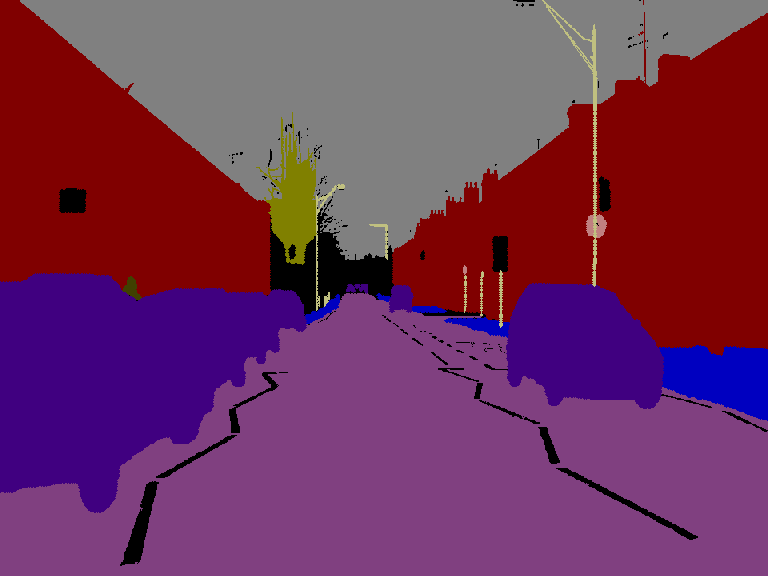}} &
      \adjustbox{valign=m,vspace=.05mm}{\includegraphics[width=0.16\textwidth]{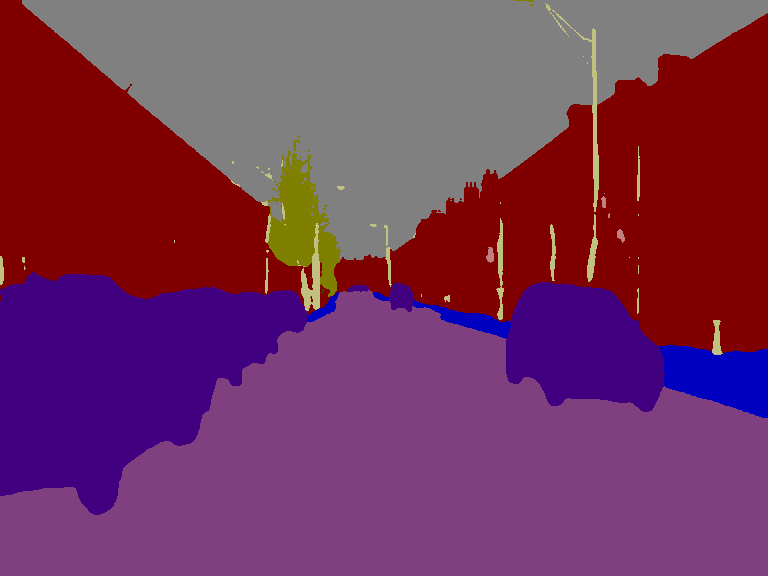}} &
      \adjustbox{valign=m,vspace=.05mm}{\includegraphics[width=0.16\textwidth]{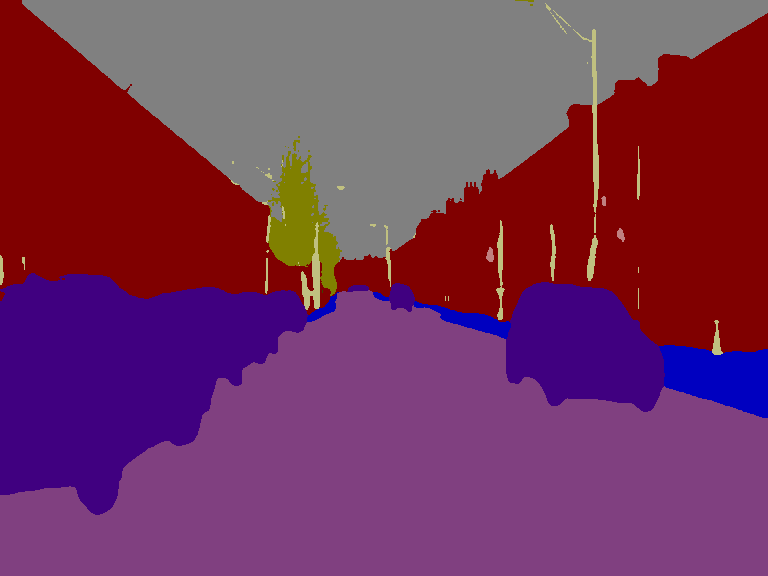}} &
      \adjustbox{valign=m,vspace=.05mm}{\includegraphics[width=0.16\textwidth]{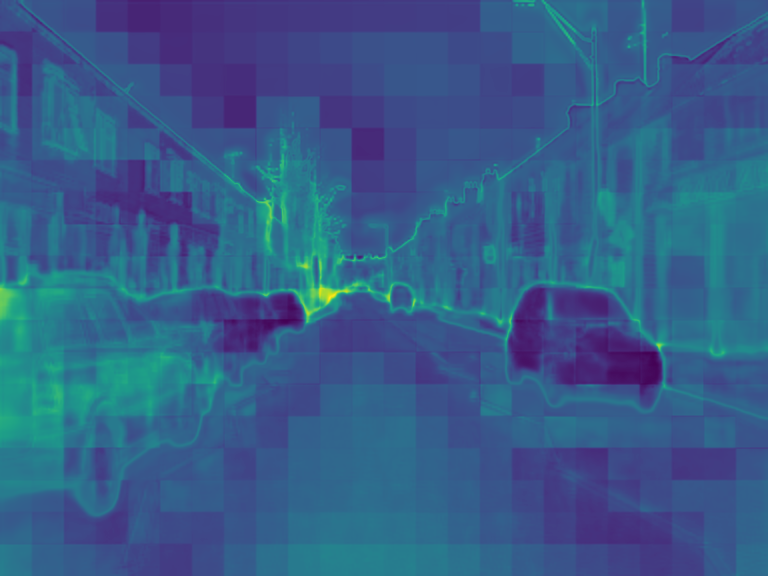}} &
      \adjustbox{valign=m,vspace=.05mm}{\includegraphics[width=0.16\textwidth]{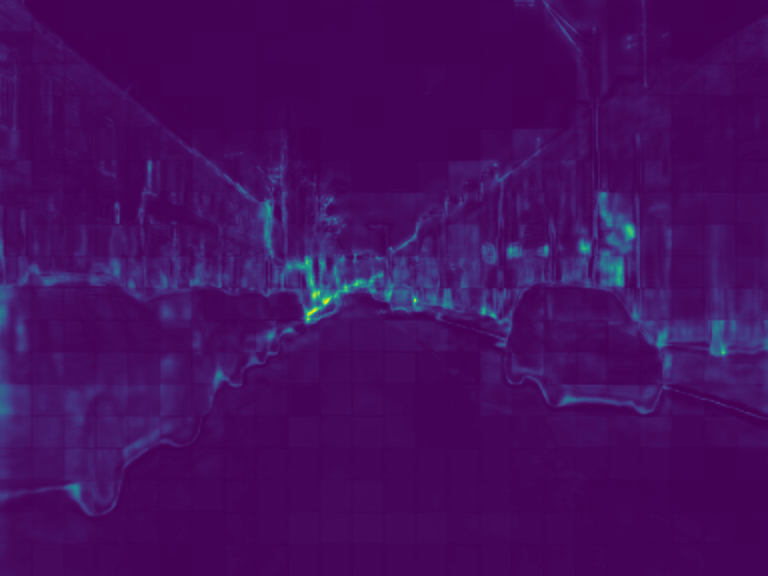}}\\[2.mm]

      \multicolumn{7}{c}{\includegraphics[width=1.\textwidth]{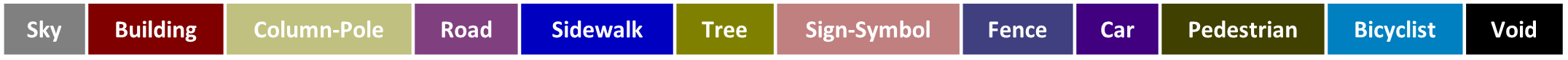}}
  \end{tabular}

  \vspace{-.5mm}
  
  \caption{Test segmentation results from the CamVid model. GT is the ground truth. MCD is the average of the results obtained with MC-dropout. For the entropy and the mutual information (MI) plots, the warmer colors represent higher values (best viewed in color)\looseness -1}
  \label{fig:camvid_example_images}

  \vspace{-2mm plus 0.5mm}
\end{figure}

Additionally to DenseNet~\cite{NIPS2017_2650d608}, we investigate the performance of HyperSeg, for which we start with $P=10\%$ images and $S=1.5$.
The model trained with active learning outperformed the model trained with random images until around 25\% of the images were queried, \cref{fig:camvid_pipeline_mi}(b). After 25\% of the images were selected, the performance of the models trained with images queried with uncertainty and randomly were very similar and stabilized around 69.0\%. A possible reason for this behavior is that HyperSeg model has excellent generalization capabilities, requiring fewer data to achieve a stable meanIoU close to the performance obtained with the entire dataset. Finally, \cref{fig:camvid_example_images} and \cref{tab:camvidResults} show results obtained with the models from the last iterations of the DenseNet and HyperSeg experiments. \Cref{tab:camvidResults} also shows the performance of the models trained with the entire datasets.

In summary, applying active learning to CamVid confirmed the effectiveness of the active learning framework. It further reinforced that when data follows a consistent pattern, adding more samples does not necessarily improve the model's performance. As shown in \cref{fig:camvid_pipeline_mi}(a-b), the meanIoU gain for DenseNet from the beginning to the end of the experiment is much more expressive than for HyperSeg. Even though more studies should be performed to analyze this behavior, we hypothesize that it regards the models' structure. HyperSeg apparently has a huge capability of generalization and does not require as much data as DenseNet to achieve the best performance possible. 

\begin{table*}[b]
\centering
\vspace{-2mm}
\caption{Results for the CamVid test subset. mIoU refers to mean IoU.}
\label{tab:camvidResults}
\renewcommand \tabcolsep {3pt}
\begin{tabular}{@{}>{\small}l*{13}{c}}
    {\rotatebox[origin=l]{0}{\footnotesize Experim.\strut}} &
    {\rotatebox[origin=c]{90}{\footnotesize \% Data}} &
    {\rotatebox[origin=c]{90}{{\footnotesize Sky}}} &
    {\rotatebox[origin=c]{90}{{\footnotesize Build.}}}   &
    {\rotatebox[origin=c]{90}{{\footnotesize Column}}} &
    {\rotatebox[origin=c]{90}{{\footnotesize Road}}} &
    {\rotatebox[origin=c]{90}{{\footnotesize Sidew.}}}  &
    {\rotatebox[origin=c]{90}{{\footnotesize Tree}}}  &
    {\rotatebox[origin=c]{90}{{\footnotesize Sign}}}   &
    {\rotatebox[origin=c]{90}{{\footnotesize Fence}}}   &
    {\rotatebox[origin=c]{90}{{\footnotesize Car}}} &
    {\rotatebox[origin=c]{90}{{\footnotesize Pedes.}}}  &
    {\rotatebox[origin=c]{90}{{\footnotesize Bicyc.}}} &
    {\rotatebox[origin=c]{90}{ {\footnotesize mIoU}}} \\
    \midrule
    \multirow{2}{*}{ \footnotesize DenseNet } &
    {\footnotesize 100}&
    {\footnotesize 88.8}&
    {\footnotesize 74.3}&
    {\footnotesize 30.8}&
    {\footnotesize 89.5}&
    {\footnotesize 77.0}&
    {\footnotesize 71.3}&
    {\footnotesize 33.1}&
    {\footnotesize 32.2}&
    {\footnotesize 68.1}&
    {\footnotesize 51.6}&
    {\footnotesize 45.6}&
    {\footnotesize 60.2}  \\
     &
    {\footnotesize41.9}&
    {\footnotesize 91.9}&
    {\footnotesize 77.6}&
    {\footnotesize 28.2}&
    {\footnotesize 92.0}&
    {\footnotesize 77.6}&
    {\footnotesize 74.1}&
    {\footnotesize 26.5}&
    {\footnotesize 26.0}&
    {\footnotesize 70.0}&
    {\footnotesize 47.5}&
    {\footnotesize 37.3}&
    {\footnotesize 59.0}\\
    \midrule
    \multirow{2}{*}{ \footnotesize HyperSeg } &
    {\footnotesize 100}&
    {\footnotesize 94.5}&
    {\footnotesize 92.9}&
    {\footnotesize 50.3}&
    {\footnotesize 97.4}&
    {\footnotesize 88.5}&
    {\footnotesize 86.4}&
    {\footnotesize 35.3}&
    {\footnotesize 70.8}&
    {\footnotesize 94.4}&
    {\footnotesize 73.8}&
    {\footnotesize 86.8}&
    {\footnotesize 79.2}\\
     &
    {\footnotesize 28.8}&
    {\footnotesize 85.8}&{\footnotesize 84.5}&{\footnotesize 38.5}&{\footnotesize 94.3}&{\footnotesize 7.2}& {\footnotesize 79.9}& {\footnotesize 47.7}&{\footnotesize 55.5}&{\footnotesize 88.7}&{\footnotesize 46.5}& {\footnotesize 61.0}&{\footnotesize 69.0}\\
\end{tabular}
\end{table*}

\begin{figure}[tp] 
    \centering
    \includegraphics[
      width = 0.7 \linewidth,
      clip,
      trim = 0mm 2.5mm 0mm 2.5mm
    ]{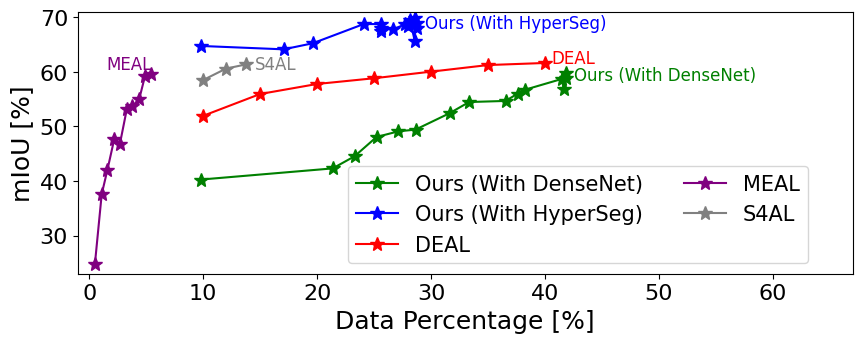}
    \vspace {-1mm}
    \caption{Comparison of methods on CamVid. The values for DEAL, MEAL, and S4AL were manually extracted from the graphs in the original papers, hence are approximate. For each framework, the markers indicate when new images were labeled, and the models were (re-)trained in the respective original papers. (Best viewed online in color.)}
    \label{fig:camvid_comparison}

    \vspace {-3mm}
\end{figure}

\Cref{fig:camvid_comparison} compares our results with three recent deep active learning methods. 
\Gls{S4AL} achieved around 61.4\% meanIoU training on 13.8\% of the data, which is 97\% of the performance obtained with the entire dataset\,\cite{rangnekar2023semantic}.
Similarly, \Gls{MEAL} achieved 59.6\% meanIoU, which is 81.6\% of the overall performance, with just 5\% of the images\,\cite{sreenivasaiah2021meal}. 
However, these frameworks queried patches of images instead of the whole image. 

\gls{DEAL} used image-level querying, and achieved 61.64\% meanIoU, which is about 95\% of the whole dataset performance, using 40.0\% of the data\,\cite{xie2020deal}. 
Using DenseNet, we achieved 97.9\% of the whole dataset's results with 41.9\% of the data. With HyperSeg, we obtained 87.1\% performance using only 28.9\% of the data.
As HyperSeg is a state-of-the-art model tailored for CamVid, the meanIoU is higher than for the other approaches.  
Notice also that the same framework requires a different percentage of data to obtain results close to the result obtained with 100\% of the data when different model architectures are used, as we demonstrated using DenseNet and HyperSeg. Finally, when analyzing the results, it is important to remember that S4AL, DEAL, and MEAL are much more complex frameworks than ours. S4AL uses a teacher-student architecture for allowing semi-supervised training using pseudo labels. MEAL uses \gls{UMAP} to learn a low dimensional embedding representation of the encoder output and uses K-Means++ to find the most representative between the most informative images sampled with entropy. DEAL adds to the CNN structure a probability attention module for learning semantic difficulty maps. Our framework consists of taking an out-of-the-box model without modifications and allowing dropout during the inference phase for using the MC-dropout and calculate the epistemic uncertainty.
\looseness -1

\paragraph{The Pipeline Dataset.}

The active learning process takes longer on this dataset, because it is larger than CamVid.
We chose to run this experiment with HyperSeg, as it required less data and fewer iterations in the CamVid experiment, while achieving significantly higher meanIoU than DenseNet.
\looseness -1

\begin{figure*}[tb]
  \centering

    \begin{tabular}{@{}c@{\hspace{0.3mm}}c@{\hspace{0.3mm}}c@{\hspace{0.3mm}}c@{\hspace{0.3mm}}c@{\hspace{0.3mm}}c@{}}

      \scriptsize Original &
      \scriptsize GT &
      \scriptsize 1 Pass &
      \scriptsize MCD &
      \scriptsize Entropy &
      \scriptsize MI \\

      \adjustbox{valign=m,vspace=.05mm}{\includegraphics[width=0.16\textwidth]{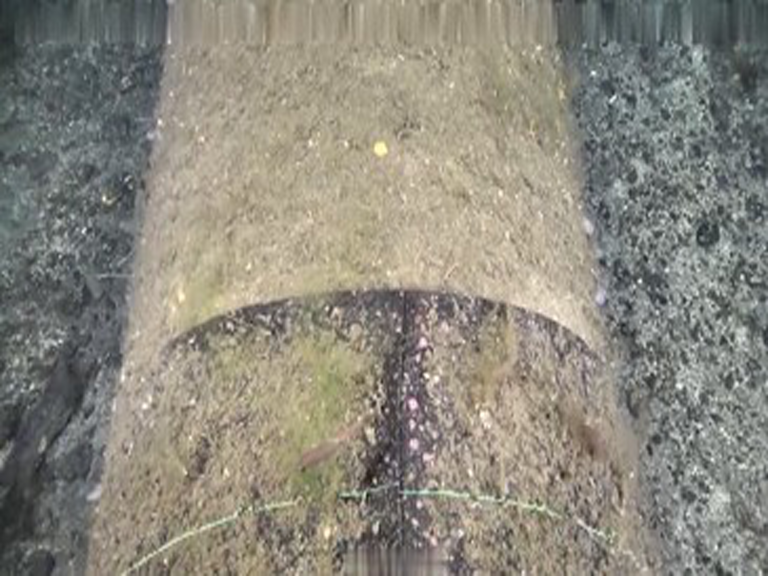}} &
      \adjustbox{valign=m,vspace=.05mm}{\includegraphics[width=0.16\textwidth]{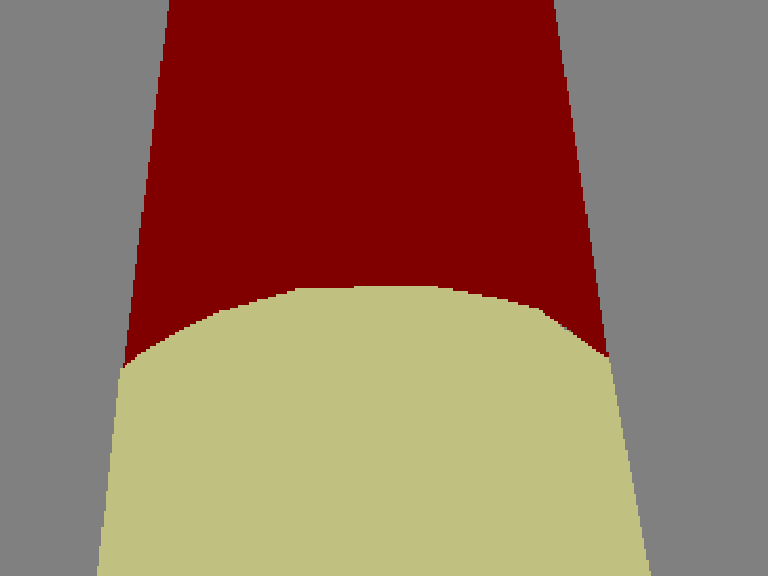}} &
      \adjustbox{valign=m,vspace=.05mm}{\includegraphics[width=0.16\textwidth]{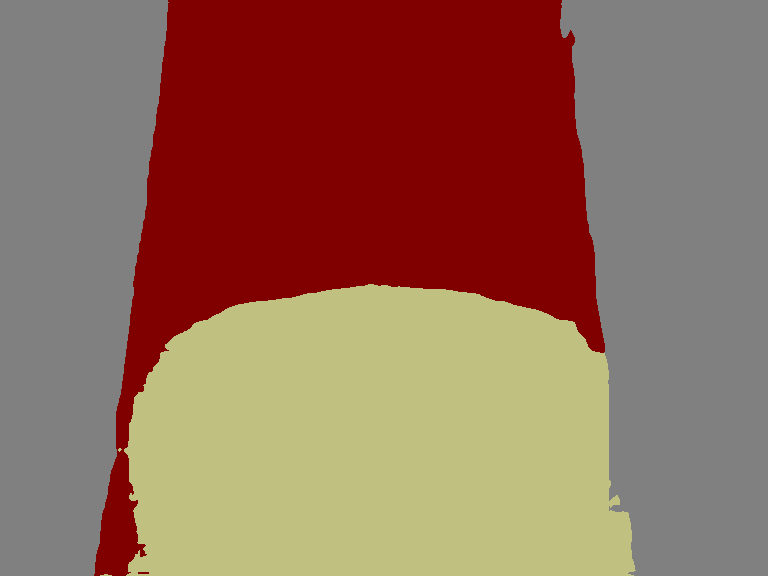}} &
      \adjustbox{valign=m,vspace=.05mm}{\includegraphics[width=0.16\textwidth]{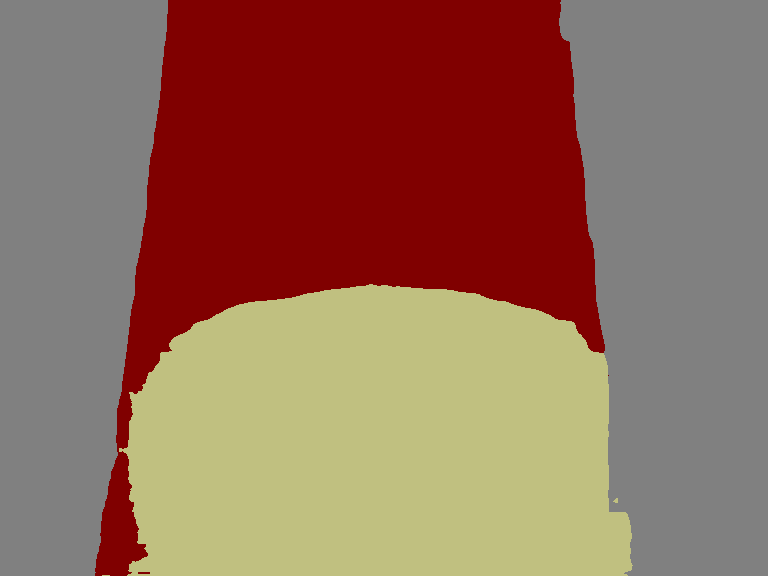}} &
      \adjustbox{valign=m,vspace=.05mm}{\includegraphics[width=0.16\textwidth]{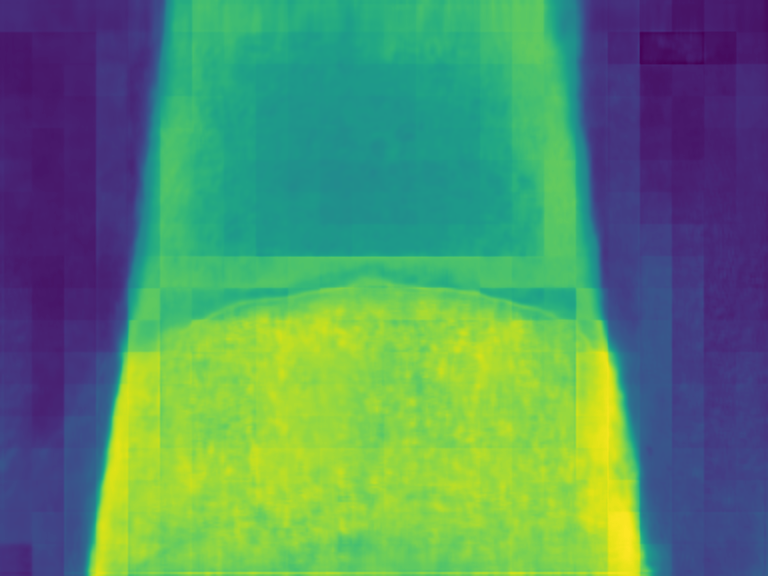}} &
      \adjustbox{valign=m,vspace=.05mm}{\includegraphics[width=0.16\textwidth]{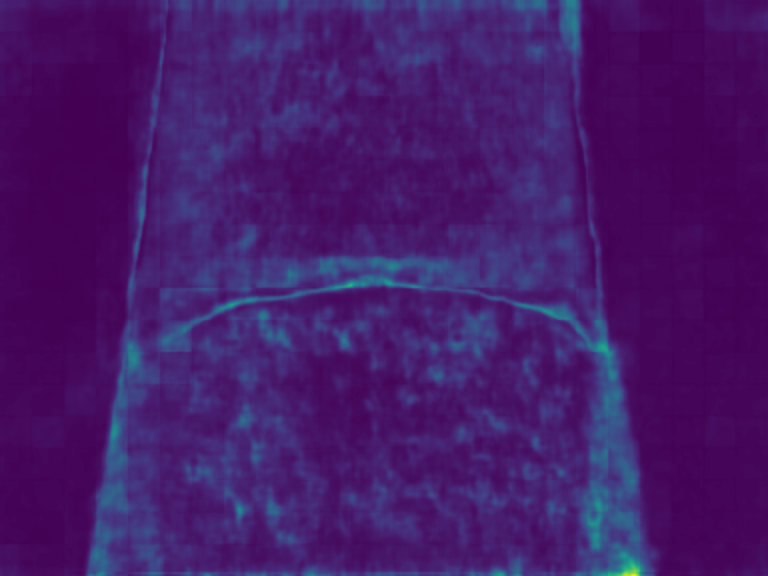}}
      \\
      [2.mm]

      \adjustbox{valign=m,vspace=.05mm}{\includegraphics[width=0.16\textwidth]{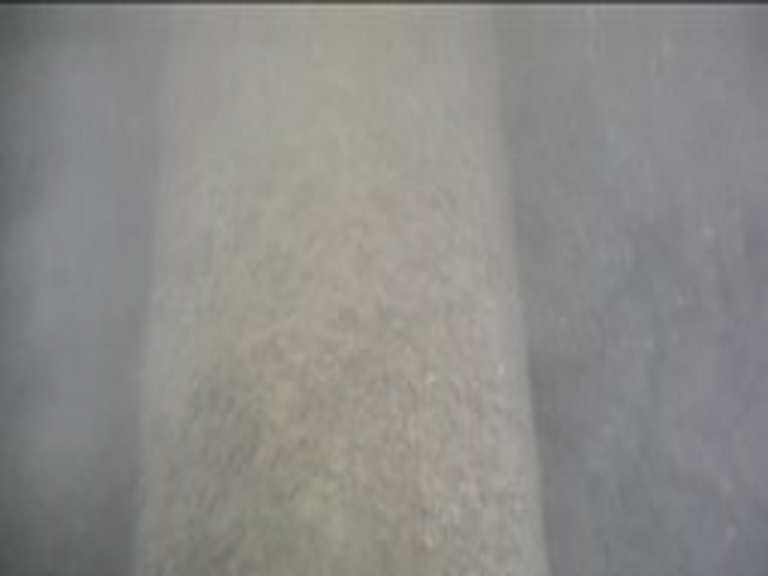}} &
      \adjustbox{valign=m,vspace=.05mm}{\includegraphics[width=0.16\textwidth]{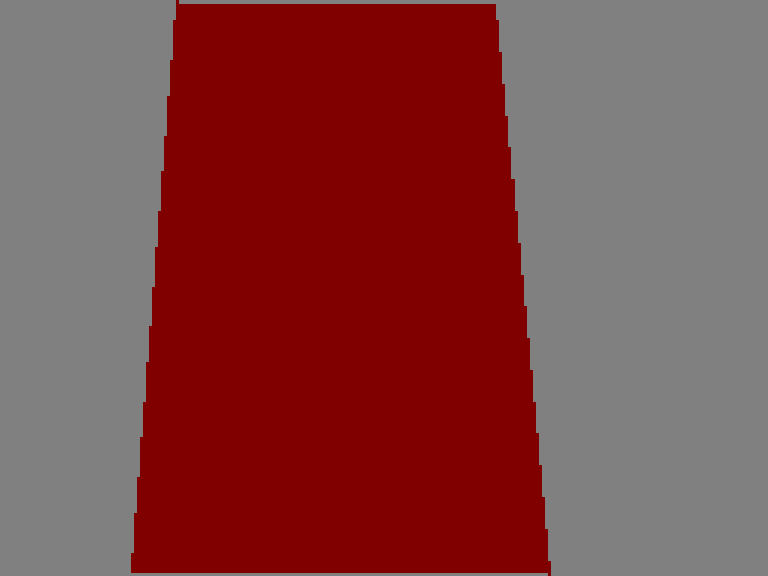}} &
      \adjustbox{valign=m,vspace=.05mm}{\includegraphics[width=0.16\textwidth]{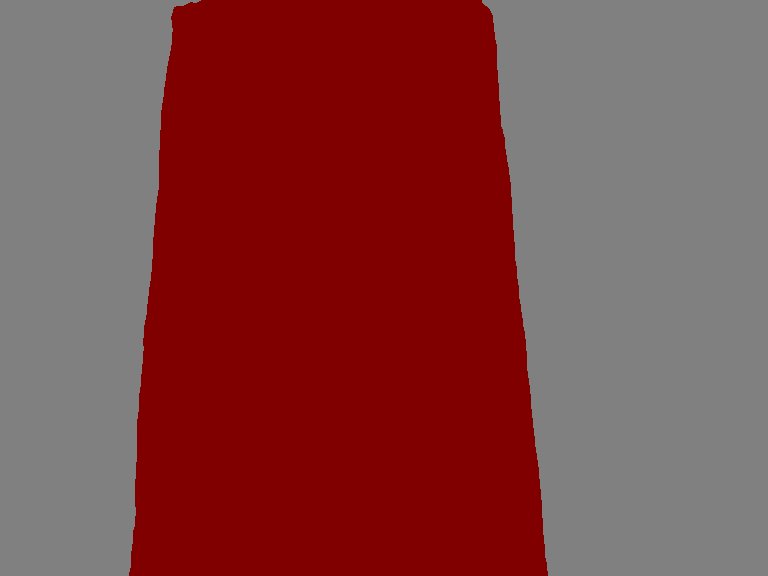}} &
      \adjustbox{valign=m,vspace=.05mm}{\includegraphics[width=0.16\textwidth]{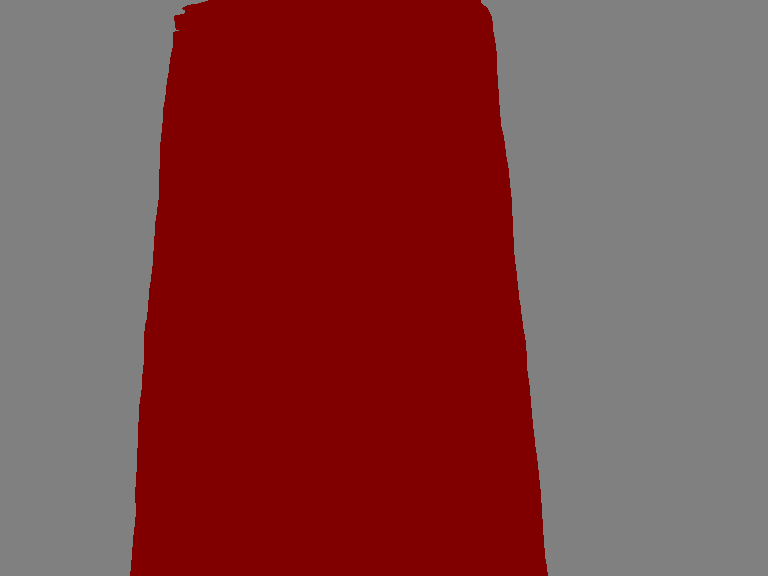}} &
      \adjustbox{valign=m,vspace=.05mm}{\includegraphics[width=0.16\textwidth]{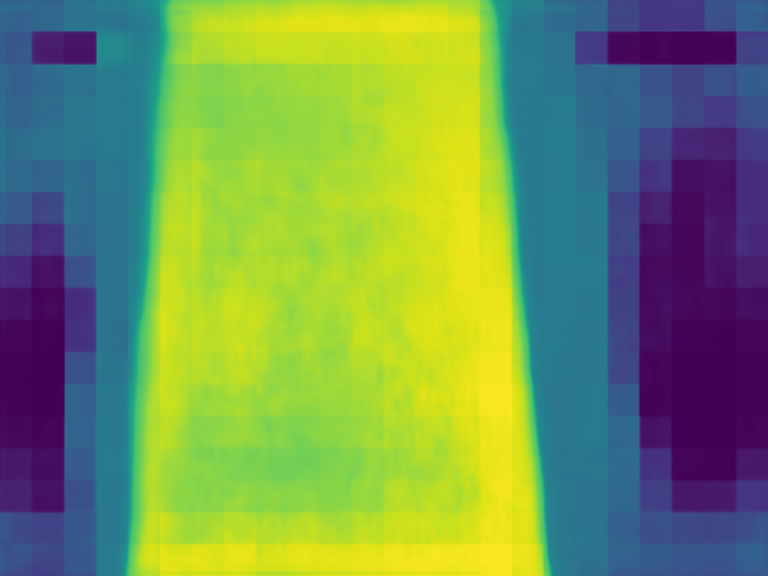}} &
      \adjustbox{valign=m,vspace=.05mm}{\includegraphics[width=0.16\textwidth]{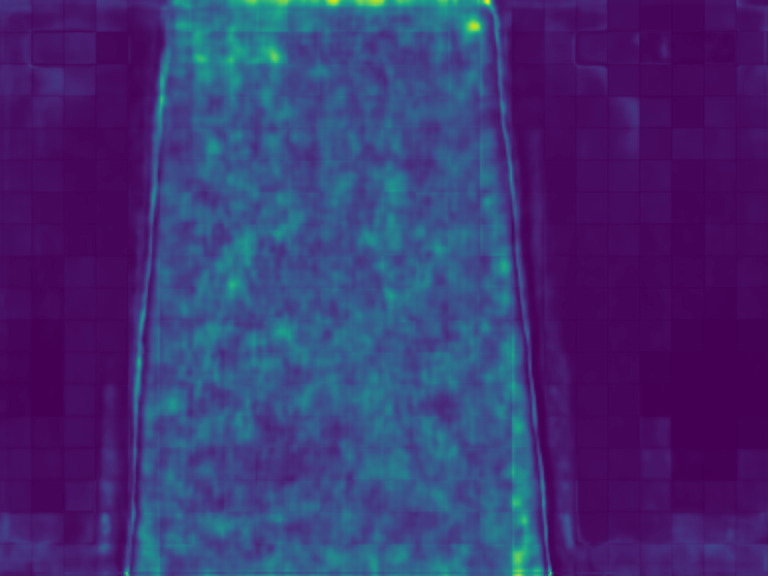}}
      \\[2.mm]
      \multicolumn{6}{c}{\includegraphics[width=0.8\textwidth]{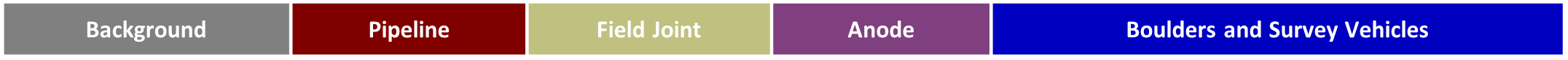}}
  \end{tabular}
  \vspace{-.7mm}
  \caption{Example results fort pipeline test images. For the entropy and the mutual information plots, the warmer colors represent higher values (best viewed in color).}
  \label{fig:pipeline_example_images}

  \vspace{-3mm plus .5mm}
\end{figure*}

\Cref{fig:camvid_pipeline_mi}(c) presents the results when starting with $P= 5\%$ of the images and $S = 1.5$. The model trained with uncertainty-based selection outperforms the baseline model trained with randomly selected images. The final meanIoU of the active learning model is 6.17\% higher than the baseline. Except for iteration\,4, the model trained with active learning was much more stable, presenting better and increasing performance across iterations. \Cref{tab:pipelineResults} presents the meanIoU for the test dataset for the final iteration models with uncertainty-based selection and with the baseline random selection. Note that for the most underrepresented class (boulder and survey vehicle)  the model trained with uncertainty-based selection presents the most significant performance gain compared to the model trained with random images.  Figure \ref{fig:pipeline_example_images} showcases two examples of predictions for test images using the resulting model from the last iteration of this experiment. The entropy plots show higher values for the pipeline and the field joint, the relatively illegible classes, suggesting that using entropy as an acquisition function could yield good results.
\looseness -1



\begin{table*}[b]
\vspace{-2mm}
\caption{mIoU for the pipeline test dataset, for the model trained in the final iteration using $12.5\%$ of the data. The percentages over headings indicate the amount of images containing each class in the entire training and validation datasets.
}
\label{tab:pipelineResults}
\vspace{-1mm plus .3mm}
\small
\renewcommand{\tabcolsep}{2mm}
\begin{tabularx}{\textwidth}{@{}Xrrrrrr@{}}
&   \scriptsize $99.9\%$ &
    \scriptsize $77.1\%$ &
    \scriptsize $20.3\%$  &
    \scriptsize $21.0\%$ &
    \scriptsize $15.5\%$ &
     \\
\footnotesize Selection Criteria &
    {\footnotesize Backg.} &
    \footnotesize Pipe &
    \footnotesize F. Joint  &
    \footnotesize Anode &
    \footnotesize B.\&S.\,V. &
    \footnotesize mIoU \\
    \midrule 
\footnotesize Uncertainty &
    {\footnotesize 97.0}&{\footnotesize 84.6}&{\footnotesize 66.3}&{\footnotesize 31.1}&{\footnotesize 58.6}&{\footnotesize 67.5}\\
\footnotesize Random (baseline) &
    {\footnotesize 96.2}&{\footnotesize 80.0}&{\footnotesize 61.3}&{\footnotesize 29.0}&{\footnotesize 40.3}&\footnotesize 61.4 \\
\end{tabularx}
\end{table*}

In \cref{fig:camvid_pipeline_mi} the difference of meanIoU between the models trained with images selected based on uncertainty and the ones trained with randomly selected images is more significant in the pipeline experiment than in the CamVid experiment. The possible reason for that is that the pipeline dataset is much bigger and unbalanced than CamVid, making the selection of images more critical and challenging.
\looseness -1

\paragraph{Repeatability.}

\begin{figure}[tp] 
  \centering
  \includegraphics[
      width = 0.97\linewidth, 
      clip,
      trim = 0mm 2.5mm 0mm 2.5mm
  ]{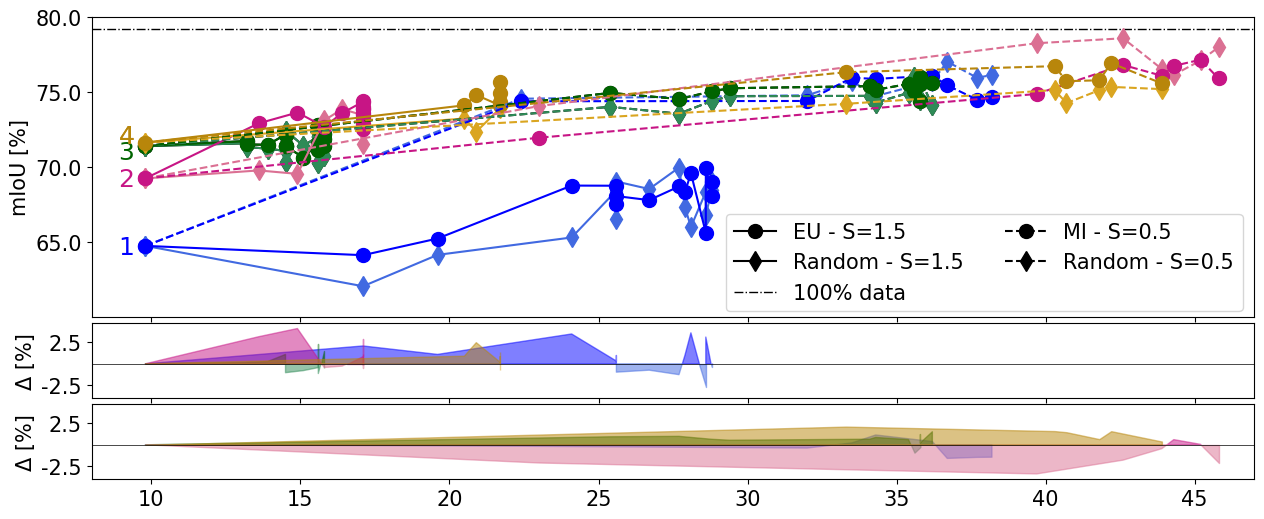}
  \vspace{-.7mm}
  \caption{
    Top: Four colors group run for different random initial sets of images. Below: mIoU differences $\mathsf\Delta$ between the active learning and the baseline random selection models for $S=1.5$ (middle) and $S=0.5$ (bottom).  Positive  \( \mathsf \Delta \) means active learning prevails. (Best viewed online in color.)\looseness -1}
    \label{fig:camvid_hyperseg_manyIterations}
  \vspace{-3mm plus .5mm}
\end{figure}

Learning depends on 
the initial set of images chosen for the first iteration. To test repeatability 
we reran the CamVid experiment with HyperSeg several times.
While the original setup for CamVid with HyperSeg used $S=1.5$ for thresholds in \cref{eq:tr}, we now used both $S=1.5$ and $S=0.5$, the latter selecting more images in each iteration. The top plot in \cref{fig:camvid_hyperseg_manyIterations}  shows mIoU results after each iteration for active learning and baseline random-selection models,  grouping runs starting with different initial sets of images using four colors.
As can be observed in the middle plot, active learning always prevails with $S=1.5$. The bottom graph shows that for $S=0.5$ the performance gain was smaller, and in the experiment number 2, in pink, random selection performaned better. We hypothesize that as CamVid is a very small dataset and HyperSeg has a strong generalization capacity, when more data is selected the benefit of the uncertainty selection gets much lower. It remains an open question whether retraining the models from scratch at each iteration would prevent them from getting stuck in local minima yielding better performance. Notice that we reported in the previous paragraphs the performance for the experiment 1, in blue. The other experiments presented better meanIoU\%. Experiment 4, in yellow, for example, achieved 75.7\% meanIoU, 95.6\% of the performance with the entire dataset, training on 21.7\% of the data. \looseness -1

\section{Conclusion}
\label{Sec:conclusion}

We demonstrated the effectiveness of active learning with epistemic uncertainty in an underwater infrastructure inspection task, using HyperSeg, a five-class dataset of more than fifty thousand images, and mutual information as the epistemic uncertainty measure. The HyperSeg structure did not need to be modified, making this method easy to implement. Using active learning for selecting the training images resulted in a model with 6.17\% better meanIoU than a baseline model trained with the same numer of random images. The model trained with active learning achieved 67.5\% meanIoU using only 12.5\% of the available data for training and validation. We observed that in the second iteration, the images queried attempted to compensate for the less represented classes in the dataset and the classes with lower performance in the previous iteration. This indicates that the approach helps with unbalanced datasets.
\looseness -1

%
%
%

%
%
%
%
\bibliographystyle{splncs04}
\bibliography{References}

\end{document}